\documentclass[runningheads]{llncs}

\usepackage{eccv}

\usepackage{eccvabbrv}

\usepackage{graphicx}
\usepackage{booktabs}

\usepackage{bbm}
\usepackage{tabularx}
\usepackage{multirow} 
\usepackage{array}
\usepackage{color}
\usepackage{algorithm,algorithmic}
\usepackage{amsmath}
\usepackage{makecell}

\usepackage[accsupp]{axessibility}  

\usepackage{hyperref}

\usepackage{orcidlink}

\begin{document}

\title{Learned Rate Control for Frame-Level Adaptive Neural Video Compression via Dynamic Neural Network} 

\titlerunning{Learned Rate Control for Adaptive NVC}

\author{Chenhao Zhang\inst{1}\orcidlink{0009-0000-2413-9064} \and
Wei Gao\inst{1,2}\thanks{Corresponding Author: Wei Gao. This work was supported by The Major Key Project of PCL (PCL2024A02), Natural Science Foundation of China (62271013, 62031013), Guangdong Province Pearl River Talent Program (2021QN020708), Guangdong Basic and Applied Basic Research Foundation (2024A1515010155), Shenzhen Science and Technology Program (JCYJ20230807120808017), and Sponsored by CAAI-MindSpore Open Fund, developed on OpenI Community (CAAIXSJLJJ-2023-MindSpore07).}\orcidlink{0000-0001-7429-5495}}

\authorrunning{C.~Zhang and W.~Gao.}

\institute{SECE, Shenzhen Graduate School, Peking University, Shenzhen, China\and Peng Cheng Laboratory, Shenzhen, China\\
\email{chenhaozhang@stu.pku.edu.cn, 
gaowei262@pku.edu.cn}\\}
\maketitle

\begin{abstract}
Neural Video Compression (NVC) has achieved remarkable performance in recent years. However, precise rate control remains a challenge due to the inherent limitations of learning-based codecs. To solve this issue, we propose a dynamic video compression framework designed for variable bitrate scenarios. First, to achieve variable bitrate implementation, we propose the Dynamic-Route Autoencoder with variable coding routes, each occupying partial computational complexity of the whole network and navigating to a distinct RD trade-off. Second, to approach the target bitrate, the Rate Control Agent estimates the bitrate of each route and adjusts the coding route of DRA at run time. To encompass a broad spectrum of variable bitrates while preserving overall RD performance, we employ the Joint-Routes Optimization strategy, achieving collaborative training of various routes. Extensive experiments on the HEVC and UVG datasets show that the proposed method achieves an average BD-Rate reduction of 14.8\% and BD-PSNR gain of 0.47dB over state-of-the-art methods while maintaining an average bitrate error of 1.66\%, achieving Rate-Distortion-Complexity Optimization (RDCO) for various bitrate and bitrate-constrained applications. Our code is available at \url{https://git.openi.org.cn/OpenAICoding/DynamicDVC}.
  \keywords{Neural Video Compression \and Rate Control \and Rate-Distortion-Complexity Optimization}
\end{abstract}

\section{Introduction}
Nowadays, video content accounts for over 80\% of internet traffic\cite{VideoInternetTraffic2022}, underscoring the critical importance of flexible video compression techniques to handle the enormous data volume. Traditional video compression standards\cite{AVC2003,HEVC2012,VVC2021} have been pivotal, incorporating technologies like inter-frame prediction, transform coding, and motion estimation. These methods employ adaptable rate control models to maximize reconstruction quality within the constraints of bandwidth or storage capacities.

Recent advancements in Neural Video Compression (NVC)\cite{AlphaVC,CANFVC2022,CANFVCPlusPlus2023,DCVC_DC2023,DCVC_HEM2022,DCVC_TCM2022,DCVC2021,DVC2019,ScaleSpaceFlow2020,Coarse2FineNVC2022,MotionPropagationNVC2023,zheng2024,wu2021} have demonstrated significant rate-distortion (RD) improvements over traditional video compression methods, primarily attributed to their enhanced non-linear transformation capabilities for better mining spatial and temporal redundancy. However, a notable limitation of NVC methods is their training for a specific RD trade-off with a fixed value of $\lambda$, which restricts their adaptability in scenarios with bitrate constraints or varying bitrate requirements.

Some works are proposed to solve this issue. These methods mainly adopt two strategies: multi-granularity quantization\cite{DCVC_DC2023,DCVC_HEM2022,BiDirectionalFlexibleRate2022} and feature modulating\cite{ModulateVariableRates2021,VariableRateNIC2021}. The former is a neural network-based strategy that facilitates adaptive quantization by varying the quantization steps across different dimensions in the latent space. This approach allows for a smooth adjustment of bitrate by manipulating the global quantization step. The latter leverages a set of scaling networks to modulate internal feature maps along channel dimension, with variable bitrate achieved by tuning hyper-parameter $\lambda$ that controls the scaling networks. Both strategies allow for variable bitrate adjustment by either quantization step scaling or feature map modulating. Nevertheless, the final RD performance performed by linear adjustment is sub-optimal when compared to end-to-end non-linear transform coding. Furthermore, they are limited in providing precise rate control due to the absence of rate estimation method, which is crucial for accurately achieving the targeted bitrate. Li \etal\cite{RateControlLearned2022} focused on developing mathematical models that correlate the output bitrate with hyper-parameter $\lambda$. However, it relies on previous experiences to iteratively refine its parameters, resulting in low bitrate accuracy, particularly in scenarios where the content of the sequence changes swiftly.

Therefore, in this paper, we introduce an end-to-end coding framework designed to achieve adaptive rate control. The main components in the framework are Dynamic-Route Autoencoder (DRA) and Rate Control Agent (RCA). The DRA incorporates a slimmable autoencoder that navigates video frames through various coding routes, each representing a subset of the overall architecture and leading to a distinct RD trade-off. These routes leverage auto-regressive coding\cite{MinnenAR2020,ELIC2022} to efficiently explore channel-wise redundancy. To enable precise rate control, we introduce the RCA, a neural network-based agent. For each frame, the agent is tasked with predicting the bitrate for each route. It selects the optimal route for the DRA to achieve the target bitrate, utilizing a sliding window algorithm for this process. Moreover, to achieve a broad spectrum of bitrate and maintain superior RD performance, we propose the Joint-Routes Optimization strategy that collaboratively trains these routes to their corresponding diverging points of the optimal RD curve. Compared to multi-granularity quantization and feature modulating, our method achieves adaptive complexity allocation that hierarchically distributes computational complexity from the lowest-bitrate route to the highest-bitrate route, enabling an efficient management of computational resources. Experiments on HEVC and UVG datasets demonstrate that the proposed method achieves an average bitrate reduction of 13.1\% over leading benchmarks, while maintaining an average bitrate error of 1.66\%, outperforming current state-of-the-art NVC rate control methods by approximately two-fold.

The contributions of our work are as follows:

\begin{itemize}
\item[$\bullet$] To the best of our knowledge, this is the first time to achieve Rate-Distortion-Complexity Optimization in the realm of NVC. Our framework dynamically allocates computational complexity while retaining superior RD performance, which is flexible in adjusting bitrates and computational resources in various bitrate applications.
\item[$\bullet$] To realize variable bitrate selection, we propose the Dynamic-Route Autoencoder that navigates the current frame to coding routes with distinct RD trade-offs. To achieve precise rate control, the proposed Rate Control Agent content-adaptively evaluates the bitrates of each route, determining the optimal route via sliding window algorithm.
\item[$\bullet$] To achieve a broad spectrum of bitrate while preserving superior joint RD performance, we propose the Joint-Routes Optimization strategy that iteratively trains each route towards its specific diverging point from the optimal RD curve by decaying the corresponding $\lambda$.  
\end{itemize}

\section{Related Works}
\subsection{Neural Video Compression}
Video compression, a critical area of study for decades, leverages core technologies like inter-frame prediction, transform coding, and motion estimation to reduce temporal and spatial redundancy. Foundational codecs such as AVC (H.264)\cite{AVC2003}, HEVC (H.265)\cite{HEVC2012}, and VVC (H.266)\cite{VVC2021} have evolved to significantly improving compression efficiency.

The rise of deep learning in multimedia has led to the advent of NVC, with pioneering frameworks like that of Lu \etal\cite{DVC2019}, which adapted traditional video compression modules into neural network counterparts, achieving comparable RD performance to HEVC. Agustsson \etal\cite{ScaleSpaceFlow2020} introduced scale-space flow for enhanced residual coding, while Lin \etal\cite{MLVC2020} utilized multiple reference frames and motion vector refinement networks to reduce residual redundancy. Moving beyond traditional residual coding, Li \etal\cite{DCVC2021,DCVC_TCM2022,DCVC_HEM2022,DCVC_DC2023} explored conditional coding to exploit temporal redundancy more effectively, demonstrating that conditional entropy $H(x_t|x_{ref})$ is no more than residual entropy\cite{OpticalFlow2020}. Shi \etal\cite{AlphaVC} proposed the pixel-to-feature motion prediction algorithm, gaining superior RD performance over VVC. Additionally, Ho \etal\cite{CANFVC2022} and Chen \etal\cite{CANFVCPlusPlus2023} introduced Conditional Augmented Normalizing Flow (CANF) for advanced conditional inter-frame coding.

\subsection{Rate Control}
Rate control is crucial in the realm of video compression to optimize video quality within bandwidth or storage limits\cite{RQModel2005,RRhoModel2010,RLambdaModel2014,yang2023,GainedNIC2021,VariableRateNIC2021,BAO2023,RateControlLearned2022}. The primary goal of rate control is to minimize distortion loss at the Group of Pictures (GoP) level under the bit constraint $R_c$ as:

\begin{align}
\min{\sum\limits_{i}^{N}R_i+\lambda_{i}D_i},s.t.\sum\limits_{i}^{N}R_i<R_c,
\label{eq:ratecontrolgoal}
\end{align}
where $N$ is the total number of frames within a GoP, $D_i$ and $R_i$ are the distortion loss and bitrate for the $i$-th frame, and $\lambda_i$ is the associated Lagrange multiplier. Traditional rate control methods such as R-$Q$\cite{RQModel2005}, R-$\rho$\cite{RRhoModel2010}, and R-$\lambda$\cite{RLambdaModel2014} models each correlates bitrate with controllable variables of video encoding to balance RD performance.

In the realm of learning-based compression, the model rigidity and complex training parameters pose challenges for variable bitrate applications. Some Neural Image Compression (NIC) methods achieve variable bitrate by applying rate allocation. Cui \etal\cite{GainedNIC2021} allocates relatively more bits to critical channels, while Song \etal\cite{VariableRateNIC2021} achieves rate allocation with a pixel-wise quality map. In NVC, Xu \etal\cite{BAO2023} utilized Semi-Amortized Variational Inference (SAVI) for iterative latent variable updating. However, the iterative adjustments of bitrate limit their applicability in real-time scenarios due to time constraints.

To achieve real-time rate control, Li \etal\cite{RateControlLearned2022} proposed the R-D-$\lambda$ model, relating bitrate constraints with compression model parameters. Despite low time-latency, the R-D-$\lambda$ model made compromise assumptions to achieve an analytical solution. Moreover, it highly depends on coding experiences, hindering its accuracy in scenarios where sequence content rapidly changes.

\subsection{Dynamic Neural Network}
Dynamic Neural Networks (DNNs) represent a versatile category of neural networks capable of adjusting their architecture\cite{Dyformer2024,AdaptiveGraphCNN2018} or parameters\cite{InternImage2023,SnakeConv2023} dynamically in response to input data. This adaptability allows DNNs to modify their depth, width, or connectivity based on the input's complexity, context, or content, offering a more flexible approach to data processing. In the realm of NIC and NVC, Yang \etal\cite{SlimCAE2021} proposed a slimmable autoencoder that transmits images to different RD trade-offs. Tao \etal\cite{AdaNIC2023} proposed a dynamic autoencoder to facilitate content-adaptive model capacity. Hu and Xu\cite{slimmabledecoder2023} leveraged a slimmable decoder to achieve adaptive decoding complexity. However, the efficient training of joint routes and the precise rate control method remain unresolved challenges.

\section{Methodology}
\subsection{Overview}
\label{sec:intro}
\begin{figure}[t!]
\centering
\includegraphics[width=120mm]{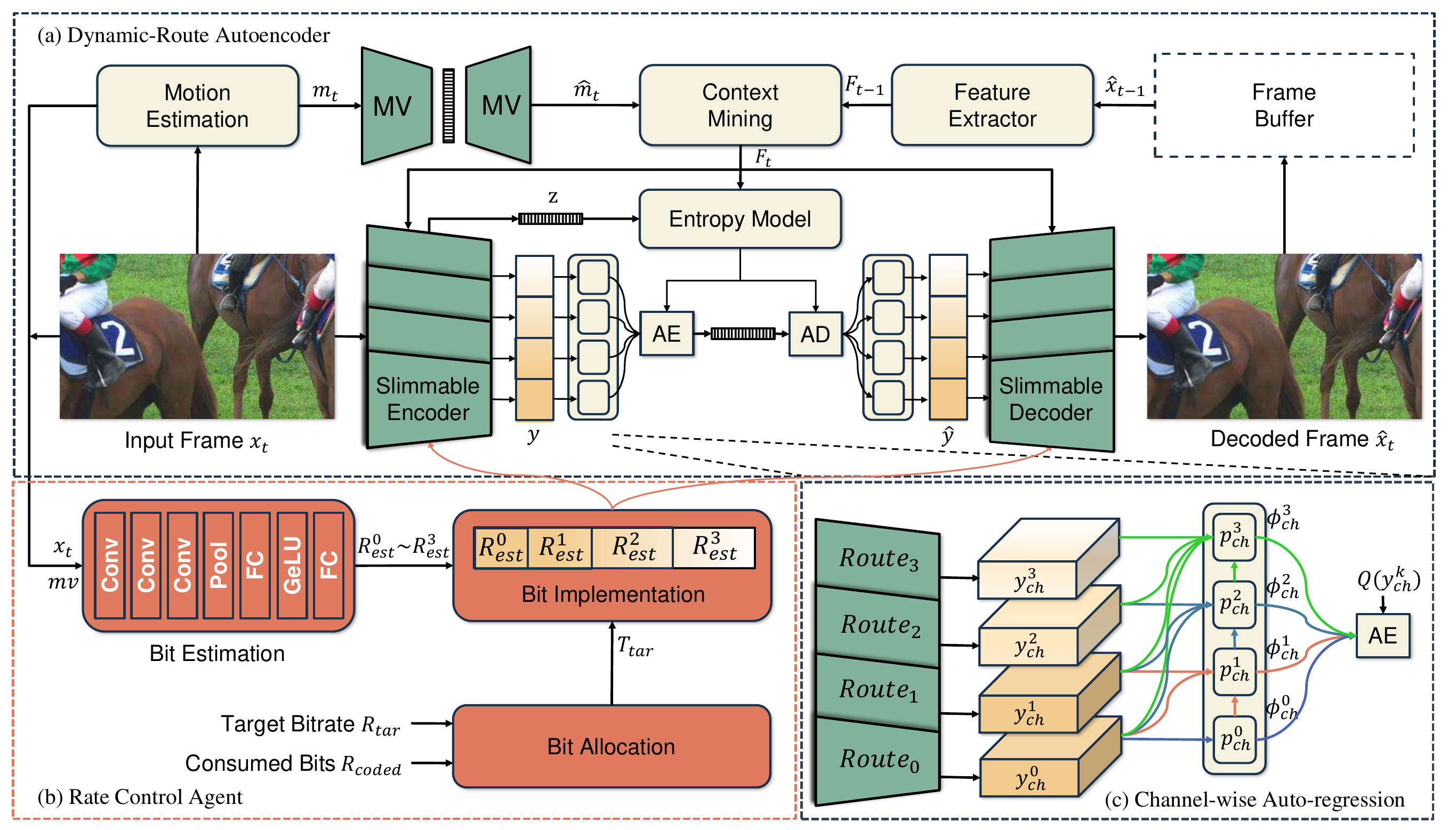}
\caption{Overview of the proposed method.}
\label{fig:framework}
\end{figure}

To achieve precise rate control while retaining optimal RD performance in a single model, we propose a dynamic NVC framework with multiple coding routes. The overall architecture is shown in \cref{fig:framework}. Specifically, RCA first estimates the bitrate of each coding route and determines the optimal route $k$ for current frame $x_t$ to approach target bitrate $R_{tar}$. Then, on the encoder side, $x_t$ is transformed into a $C_k$-channel latent representation $y_{0:C_k} \in \mathbb{R}^{H \times W \times C_k}$ via route $k$ with the help of temporal context $x_{ref}$. After that, $y_{0:C_k}$ is entropy encoded to bitstream via channel-wise auto-regression. On the decoder side, the bitstream is entropy decoded to latent representation $\hat{y}_{0:C_k}$ following the same auto-regressive procedure as entropy encoding. Finally, decoded frame $\hat{x}_t$ is reconstructed by the decoder via the same route $k$. To facilitate a broad bitrate spectrum while achieving global optimal RD performance, the coding routes are jointly trained by Joint-Routes Optimization strategy.



\subsection{Dynamic-Route Autoencoder}
\label{sec:DRA}

\begin{figure}[t!]
\centering
\includegraphics[width=100mm]{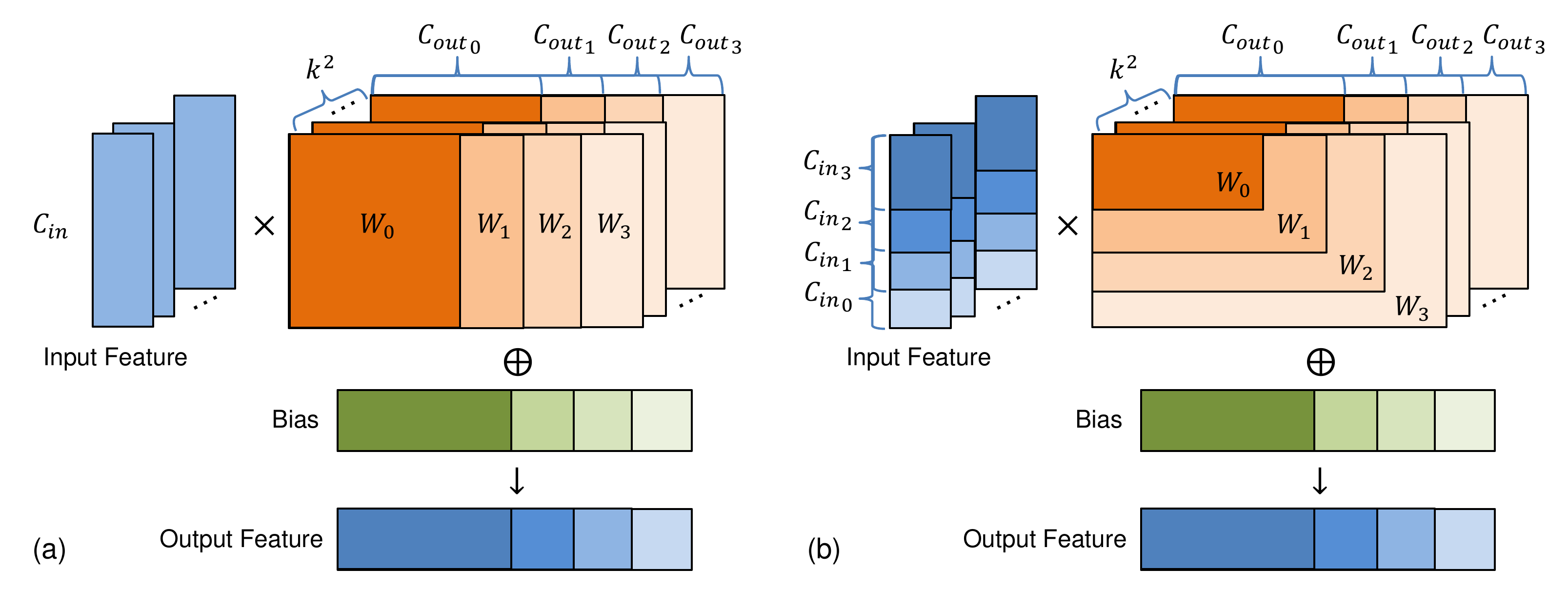}
\caption{Illustration of Slimmable Convolution Operators: (a) Conversion from fixed input channels $C_{in}$ to variable output channels $\{C_{out0}, C_{out1}, C_{out2}, C_{out3}\}$; (b) Conversion from variable input channels $\{C_{in0}, C_{in1}, C_{in2}, C_{in3}\}$ to variable output channels $\{C_{out0}, C_{out1}, C_{out2}, C_{out3}\}$}.
\label{fig:slimconv}
\end{figure}

The fundamental components of DRA are slimmable operators, which transfer input feature into output feature with dynamic channels, as shown in \cref{fig:slimconv}. Slimmable operators allow for dynamic adaptation of the \textit{supernet} into a series of overlapping sub-networks, or \textit{subnets}. Each subnet embodies a distinct, end-to-end coding route, where slimmable operators facilitate the transformation of a singular input frame into various latent representations. The adaptability in channel number empowers the framework with the capacity for variable computational complexity and rate selection. For route $k$, the coding pipeline can be formulated as follows:
\begin{align}
y_{ch}^{\leq k} &= g_{a}(x_t|x_{ref};\theta^{\leq k}), \label{eq:1} \\
\Phi_{ch}^{k} =& p_{ch}(Q(FM(y_{ch}^{\leq k-1})),\hat{z}),\label{eq:2} \\
\hat{x}_t &=g_{s}(\hat{y}_{ch}^{\leq k};\phi^{\leq k}),\label{eq:3}
\end{align}
where \(g_a(\cdot;\theta)\) and \(g_s(\cdot;\phi)\) represent the contextual encoder and decoder equipped with slimmable parameters \(\theta\) and \(\phi\), respectively, \(p_{ch}(\cdot)\) denotes the channel-wise auto-regression model that predicts entropy parameters \(\Phi_{ch}\) with the hyperprior \(\hat{z}\), \(Q(\cdot)\) refers to the quantization step, and \(FM(\cdot)\) is the Feature Modulation network. The current input frame $x_{t}$ first passes through slimmable encoder $g_a(\cdot;\theta^{\leq k})$ with parameters $\theta^{\leq k} = \{\theta^{0}, \theta^{1},...,\theta^{k}\}$, yielding latent representation $y_{ch}^{\leq k} = \{y_{ch}^{0},y_{ch}^{1},...,y_{ch}^{k}\}$, encapsulating frame information in a more condensed format. To fully explore channel redundancy, entropy parameters $\Phi_{ch}^{k}$ of $y_{ch}^{\leq k}$ are auto-regressively derived from the preceding route's output via auto-regression model $p_{ch}$, as shown in \cref{fig:framework}(c). To avoid the high coding latency caused by switching routes in auto-regression, we directly utilize the partial latent representation of the current route, $y_{ch}^{\leq k-1}$, as a replacement for the output of the previous route $k-1$. This approach avoids the need for re-encoding and streamlines the coding process. However, this substitution leads to a sub-optimal RD performance due to the absence of serial-routes optimization. To address this problem, we propose the Feature Modulation network $FM(\cdot)$ that refines $y_{ch}^{\leq k-1}$ to approximate the output of route $k-1$ before quantization and auto-regression.

\begin{figure}[t!]
\centering
\includegraphics[width=120mm]{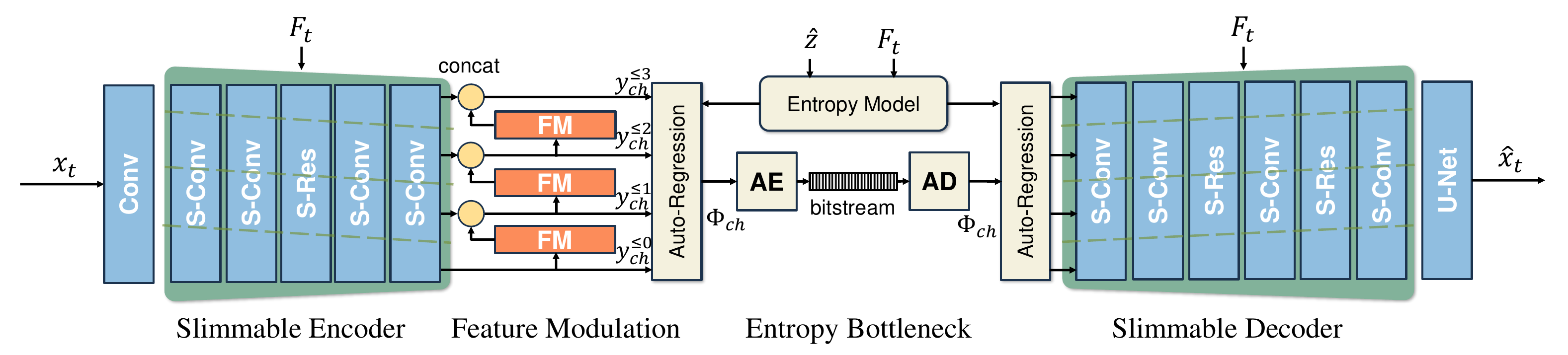}
\caption{Illustrations of slimmable autoencoder, including Slimmable Encoder, Feature Modulation, Entropy Bottelneck, and Slimmable Decoder. "S-X" denotes "Slimmable X".}
\label{fig:DRADetails}
\end{figure}

Through entropy encoding, the current frame is transferred into a bitstream for storage or transmission. The entropy decoding procedure is consistent with that of entropy encoding, which auto-regressively restores the latent representation $\hat{y}_{ch}^{\leq k}$ via $p_{ch}$. Finally, the decoded frame is reconstructed by slimmable decoder $g_s(\cdot;\phi^{\leq k})$ through the same route $k$. Implementation details of the slimmable autoencoder are shown in \cref{fig:DRADetails}.

Different from other variable rate solutions that maintain stable coding complexity at each RD point, the proposed method is complexity-adaptive, allocating less computation resources to lower-bitrate routes. This reflects the insight that simpler transformations are sufficient for lower-bitrate representations to explore source correlation, enabling effective Rate-Distortion-Complexity Optimization in our framework.

\subsection{Rate Control Agent}
As shown in \cref{fig:framework}(b), the RCA is comprised of three foundational components: the Rate Estimation module, the Bit Allocation module, and the Bit Implementation module. Given the fact that the complexity of frame content significantly influences coded bitrates, with less complex frames requiring fewer coding bits and vice versa, we first develop a lightweight Rate Estimation module, which is designed to correlate the content of current frame $x_t$ with its coded bitrates across various routes. To be specific, we leverage $x_t$ and its associated reference motion vector ($mv$) to estimate the coded bitrates $R_{est}^i$ of each route $i$. The final layer of the Rate Estimation module integrates a fully connected (FC) layer with the number of prediction heads equal to the route numbers.

For frame-level bitrate allocation, the Bit Allocation module employs the sliding window algorithm. Given a target bitrate $R_{tar}$, the bitrate of current frame is allocated as follows:
\begin{align}
T_{tar} = \frac{R_{tar}\times(N_{coded}+SW)-R_{coded}}{SW},
\label{eq:4}
\end{align}
where $T_{tar}$ is the target bitrate allocated to the current frame, $N_{coded}$ denotes the number of frames already coded, $R_{coded}$ is the bits consumed by already coded frames and $SW$ is the sliding window's length, facilitating a smoother rate control process. Guided by \cref{eq:4}, the target bitrate is expected to be achieved within $SW$ frames, ensuring a steady bitrate allocation and consistent video quality. To mitigate potential cumulative errors, the initial frame of a GoP, known as the I frame, is exempt from this bit allocation strategy and receives a relatively higher bitrate allocation to foster improved RD performance in subsequent frames.

Finally, the Bit Implementation module determines the optimal coding route $i^*$ based on the relationship between bits allocated $R_{tar} \times N_{coded}$ and bits consumed $R_{coded}$. If more bits are allocated than consumed, it chooses the route with an estimated bitrate just above $T_{tar}$, defaulting to the highest-bitrate route if none match. Conversely, if fewer bits are allocated than consumed, it chooses the route with an estimated bitrate just below $T_{tar}$, or the lowest-bitrate route if none route is available, expressed as:

\begin{equation}
i^* = 
\begin{cases} 
\mathop{\arg\min}\limits_{i}R_{est}^i - T_{tar},s.t.R_{est}^i > T_{tar}& \text{if } R_{tar} \times N_{coded} > R_{coded} \\
\mathop{\arg\min}\limits_{i}T_{tar} - R_{est}^i,s.t.R_{est}^i < T_{tar}& \text{if } R_{tar} \times N_{coded} < R_{coded}
\end{cases}
\label{eq:6}
\end{equation}

\subsection{Joint-Routes Optimization Strategy}
\label{subsec:bitallocation}
\label{sec:JROS}
NVC's primary aim is to identify the optimal encoding and decoding parameters that minimize the RD loss. The scenario becomes significantly more complex when it comes to dynamic-route NVC optimization due to the presence of multiple RD losses corresponding to various coding routes:
\begin{align}
\theta^{*}, \phi^{*}\leftarrow \mathop{\arg\min}\limits_{\theta, \phi}\sum\limits_{i=0}^{K-1}R_{i}+\lambda_{i} D_{i},
\label{eq:7}
\end{align}
where $\theta^{*}$ and $\phi^{*}$ denote the optimal parameters for the encoder and decoder, respectively, $R_{i}$ and $D_{i}$ signify the rate and distortion losses of the route $i$, $K$ is the total number of routes, and $\lambda_{i}$ is the Lagrange multiplier for route $i$. Addressing this optimization is difficult due to the parameter sharing across routes and the complexity of selecting the optimal $\lambda_{i}$ for each route. An initial training strategy is detailed in \cref{alg:naivetraining}.

\begin{algorithm}[!ht]
    \renewcommand{\algorithmicrequire}{\textbf{Input:}}
	\renewcommand{\algorithmicensure}{\textbf{Output:}}
	\caption{Initial Training Strategy}
    \label{alg:naivetraining}
    \begin{algorithmic}[1] 
        \REQUIRE training iteration $N$, number of routes $K$, $\lambda$ list $[\lambda_0,\cdots,\lambda_{K-1}]$, encoder $Enc(\cdot;\theta)$, decoder $Dec(\cdot;\phi)$, train dataset $\chi_{train}$; 
	    \ENSURE optimal coding parameters $\theta^{*}$, $\phi^{*}$; 
        \FOR {$k=0,2,\cdots,K-1$}
            \FOR {$j=1,2,\cdots,N$}
                \STATE $I_{input},I_{ref} \leftarrow \chi_{train}$
                \STATE $R, D \leftarrow Dec(Enc(I_{input}, I_{ref};\theta_k);\phi_k)$;
                \STATE $Loss \leftarrow \sum\limits_{i=0}^{K-1}R + \lambda_iD$;
                \STATE update $\theta_k, \phi_k$;
            \ENDFOR
        \ENDFOR
        \STATE \textbf{return} $\theta_{K-1}, \phi_{K-1}$.
    \end{algorithmic}
\end{algorithm}

\begin{figure}[t!]
\centering
\includegraphics[width=80mm]{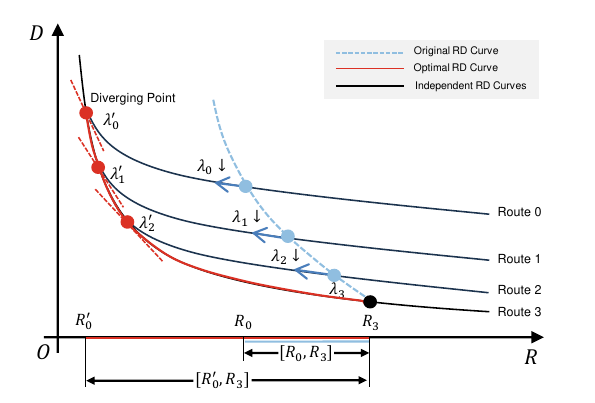}
\caption{Illustration of the diverged RD model and the Joint-Routes Optimization (JRO) strategy. The strategy focuses on transferring RD points (blue points) of each route to their corresponding diverging points (red points), shifting the total RD curve from the blue curve to the red curve. This transition broadens the bitrate spectrum and ensures global-optimal RD performance.}
\label{fig:JRO}
\end{figure}

For routes trained under fixed $\lambda$ values without parameter sharing, \cref{alg:naivetraining} offers an optimal training strategy. However, when it comes to joint routes with parameter sharing, pre-set fixed $\lambda$ values result in local-optimal instead of global-optimal RD performance, while hierarchical parameter sharing creates a cascading effect on the performance of higher-bitrate route due to adjustments in lower-bitrate routes. Addressing this, we introduce a diverged RD model that reflects the variation in RD curve behaviors across routes, as shown in \cref{fig:JRO}. This model leverages the divergence in RD curves at different bitrates, where lower-bitrate routes diverge earlier. By iteratively fine-tuning $\lambda$ values for lower-bitrate routes to align with higher-bitrate RD curves, we achieve global-optimal RD performance across routes.

To advance the initial training method, we introduce the JRO strategy, outlined in \cref{alg:jointroutesoptimization}. This strategy fine-tunes \(\lambda_i\) for each route towards its diverging point, guided by a decay coefficient \(\kappa\). A diverging point is identified when the decline in the slope of adjacent RD points ceases. To mitigate the cascading effect from parameter sharing, adjustments start with the highest-bitrate route and proceed to the lowest. By setting a higher \(\lambda_{K-1}\) for pre-training and a lower \(\lambda_{0}\) for post-training, we ensure coverage across a wide bitrate spectrum.

\begin{algorithm}[!ht]
    \renewcommand{\algorithmicrequire}{\textbf{Input:}}
	\renewcommand{\algorithmicensure}{\textbf{Output:}}
	\caption{Joint-Routes Optimization Strategy}
    \label{alg:jointroutesoptimization}
    \begin{algorithmic}[1] 
        \REQUIRE  training iteration $N$, number of routes $K$, $\lambda$ of highest route $\lambda_{K-1}$, decay coefficient $\kappa$, encoder $Enc(\cdot;\theta)$, decoder $Dec(\cdot;\phi)$, train dataset $\chi_{train}$, validation dataset $\chi_{val}$; 
	    \ENSURE optimal coding parameters $\theta^{*}$, $\phi^{*}$; 
        \STATE pre-train $\theta_{K-1}, \phi_{K-1}$ under $\lambda_{K-1}$
        \STATE test $\xi_{cur}\leftarrow\frac{D_{K-1} - D_{K-2}}{R_{K-1} - R_{k-2}}$ on $\chi_{val}$
        \STATE test $R_{cur}$ on $\chi_{val}$
        \STATE $\xi_{pre}, R_{pre}\leftarrow0$
        \FOR {$k=K-2,K-3\cdots,0$}
            \REPEAT
            \STATE $[\lambda_0:\lambda_{k+1}]\leftarrow\kappa[\lambda_0:\lambda_{k+1}]$
            \STATE $\xi_{pre}\leftarrow\xi_{cur}$  
                \FOR {$j=1,2,\cdots,N$}
                    \STATE $I_{input},I_{ref} \leftarrow \chi_{train}$
                    \STATE $R, D \leftarrow Dec(Enc(I_{input}, I_{ref};\theta_k);\phi_k)$;
                    \STATE $Loss \leftarrow \sum\limits_{i=0}^{K-1}R + \lambda_iD$;
                    \STATE update $\theta_k, \phi_k$;
                \ENDFOR
                \STATE test $\xi = \frac{D_k - D_{k+1}}{R_k - R_{k+1}}$ on $\chi_{val}$
            \STATE $\xi_{cur}, R_{cur}\leftarrow\xi,R$  
            \UNTIL {$\xi<\xi_{pre}$ and $R<R_{pre}$}
        \ENDFOR
        \STATE post-train by decaying $\lambda_0$
        \STATE \textbf{return} $\theta_{K-1}, \phi_{K-1}$.
    \end{algorithmic}
\end{algorithm}



\section{Experiments}
\subsection{Implementation Details}
\noindent\textbf{Datasets}. We train our model on the Vimeo-90k dataset\cite{Vimeo90k}, which comprises 89,800 video clips. To ensure compatibility with the autoencoder framework, sequences are randomly cropped into patches of size \(256 \times 256\). For evaluation, we test the performance of our algorithm on the HEVC standard test sequences\cite{HEVC} (Class B, C, D, E) and the UVG dataset\cite{UVG}.

\noindent\textbf{Network Implementation}. In our implementation, we set the number of routes, \(K=4\), to achieve accurate rate control across all frames. Increasing the number of routes could further improve rate control precision but at the cost of increased coding latency due to auto-regressive coding. The length of the sliding window, \(SW\), is set to 30. The number of output channels for the Motion Vector compression network and the frame compression network are set to 64 and 96, respectively. The output channel numbers for each coding route are configured as \((24,48,72,96)\).

\noindent\textbf{Evaluation Metrics}. The reconstruction quality is assessed using PSNR, while RD performance of different methods is compared using BD-Rate and BD-PSNR\cite{Bjntegaard2001}. Rate control accuracy is evaluated by bitrate error \(\Delta R=\frac{|R_{out}-R_{tar}|}{R_{tar}}\times 100\%\), and coding efficiency is measured by coding time of each frame \(T=\frac{T_{total}}{N_{frame}}\).


\noindent\textbf{Benchmark Models}. To benchmark our algorithm's RD performance and rate control accuracy, we compare the RD performance against notable NVC algorithms (\textit{i.e.,} DCVC\cite{DCVC2021}, DCVC-HEM\cite{DCVC_HEM2022}, DCVC-TCM\cite{DCVC_TCM2022}, CANF-VC\cite{CANFVC2022} and AlphaVC\cite{AlphaVC}) and the rate control accuracy against the state-of-the-art R-D-$\lambda$ model\cite{RateControlLearned2022}. To ensure fairness in comparison, the number of maximum output channels of each model is set to be consistent.

\begin{figure}[t!]
\centering
\includegraphics[width=120mm]{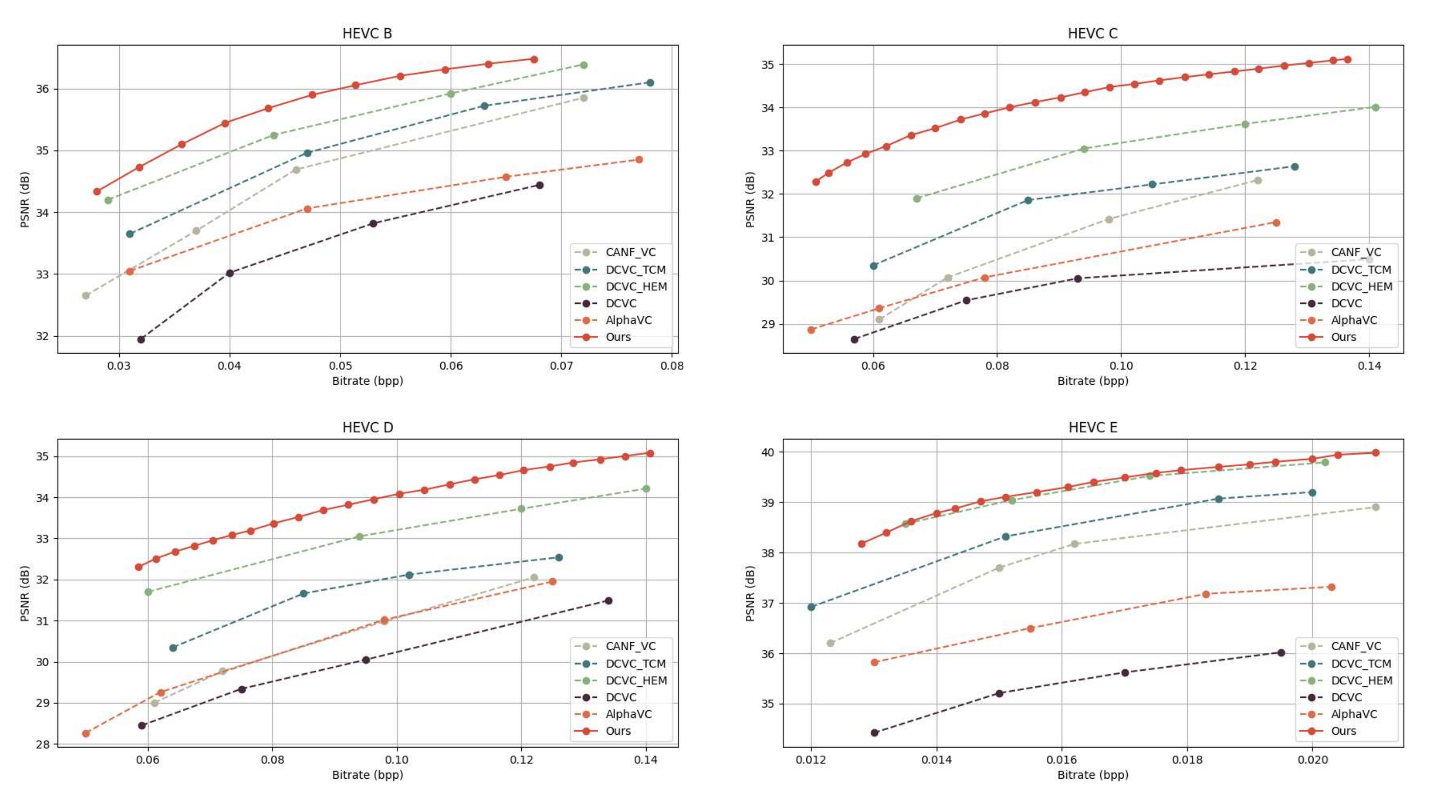}
\caption{RD performance comparison of different methods on HEVC datasets.}
\label{fig:HEVCRDPerformance}
\end{figure}

\begin{figure}[t!]
\centering
\includegraphics[width=90mm]{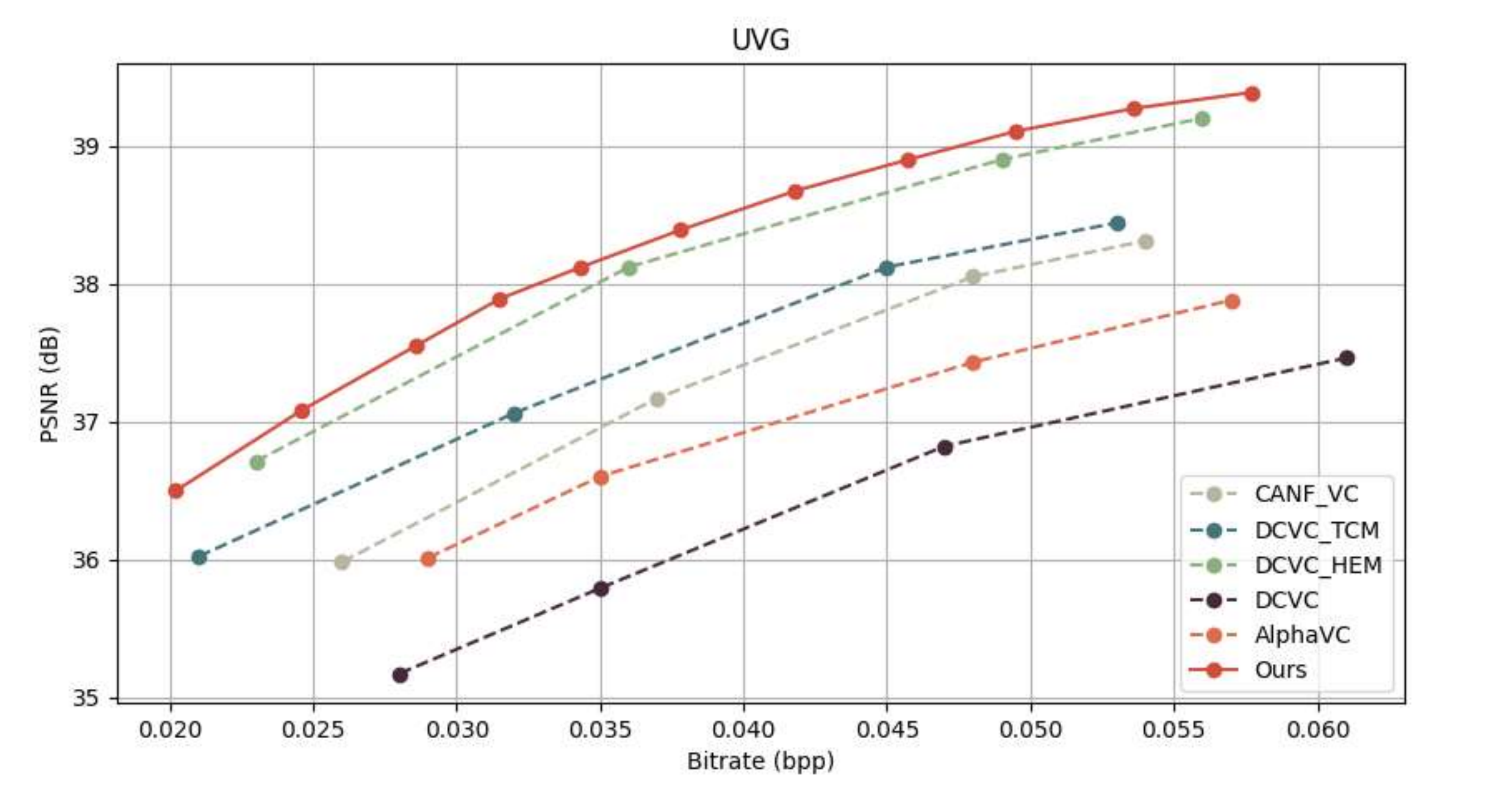}
\caption{RD performance comparison of different methods on UVG datasets.}
\label{fig:UVGRDPerformance}
\end{figure}

\subsection{Experimental Results}

\noindent\textbf{RD Performance.} \cref{fig:HEVCRDPerformance} and \cref{fig:UVGRDPerformance} illustrate the comprehensive RD performance of our proposed method. Notably, our approach achieves superior RD performance compared to the benchmark methods. Quantitative assessments, including BD-Rate comparisons, are detailed in Table \ref{tab:RDperformance}. On average, our method facilitates an average BD-Rate reduction of 14.8\% and BD-PSNR gain of 0.47dB compared to DCVC-HEM, underscoring the effectiveness of our optimization strategy in enhancing video compression efficiency.

\begin{table}[t!]
  \caption{RD performance comparison of different methods on HEVC and UVG datasets.}
  \label{tab:RDperformance}
  \centering
  \newcolumntype{C}{>{\centering\arraybackslash}X}
  \begin{tabularx}{\textwidth}{@{} p{0.105\textwidth} *{6}{C} @{}} 
    \toprule
    \multirow{2}{*}{Datasets}  & \multicolumn{5}{c}{BD-Rate(\%)/BD-PSNR(dB)} \\
    \cline{2-6}
    &  \scriptsize DCVC\cite{DCVC2021} & \scriptsize DCVC-HEM\cite{DCVC_HEM2022} &  \scriptsize DCVC-TCM\cite{DCVC_TCM2022} & \scriptsize CANF-VC\cite{CANFVC2022} & \scriptsize alphaVC\cite{AlphaVC}\\
    \midrule
    HEVC B & -57.03/2.36 & -14.509/0.37 & -28.718/0.87 & -32.258/1.18 & -53.295/1.75\\
    HEVC C & -98.656/4.27 & -36.031/1.2 & -56.723/2.31 & -58.767/3.22 & -74.38/3.68\\
    HEVC D & -59.914/3.81 & -24.641/0.84 & -52.477/2.04 & -53.736/3.08 & -57.022/3.04\\
    HEVC E & -77.665/3.88 & -5.137/0.13 & -16.414/0.69 & -29.007/1.13 & -88.706/2.59\\
    UVG  & -53.555/2.36 & -2.826/0.06 & -21.842/0.88 & -30.533/1.18 & -43.795/1.59\\
    Average  & -66.700/3.11 & -14.808/0.47 &-33.184/1.27&-39.014/1.80&-59.443/2.34\\
    \bottomrule
  \end{tabularx}
\end{table}

\begin{figure}[t!]
\centering
\includegraphics[width=120mm]{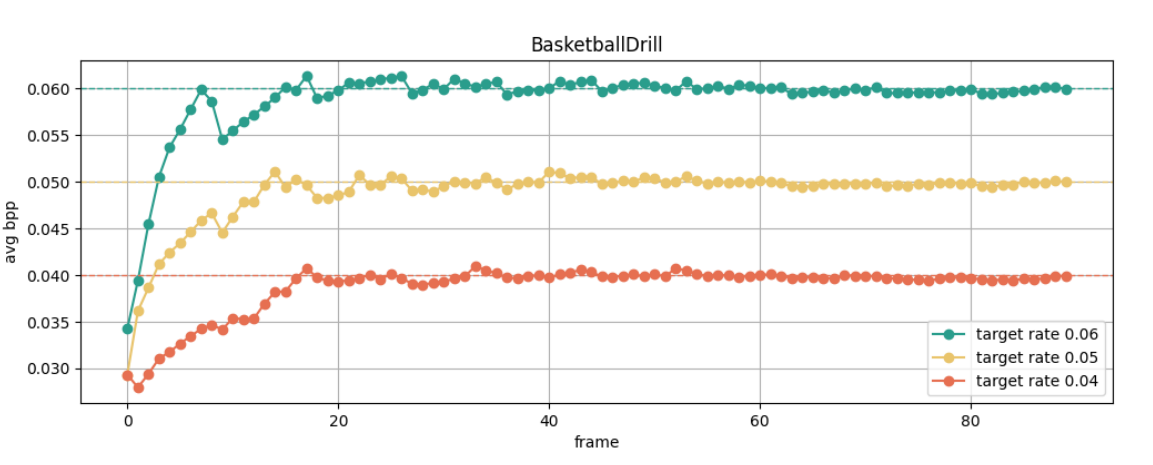}
\caption{Bitrate accuracy of proposed method with different target rates on \textit{BasketballDrill} sequence}
\label{fig:targetrate}
\end{figure}

\noindent\textbf{Rate Control Accuracy.}
Figure \ref{fig:targetrate} demonstrates the rate control performance of our proposed method. In the case of specific test sequences (\textit{BasketballDrill} as an example), the average bpp aligns with the target bpp within the span of $SW$ frames. This performance is contrasted with the state-of-the-art R-D-$\lambda$ model to highlight improvements in rate control accuracy. The quantitative results, presented in Table \ref{tab:ratecontrol}, affirm our method's capability to adhere to any specified rate within its variable range, achieving an average bitrate error of $\Delta R=1.66\%$. This level of precision significantly surpasses that of the R-D-$\lambda$ model, illustrating our approach's enhanced reliability in rate control.

\begin{table}[t!]
  \caption{Rate control accuracy comparison of the proposed method and R-D-$\lambda$ model.
  }
  \label{tab:ratecontrol}
  \centering
  \begin{tabular}{p{2cm}p{4cm}p{4cm}}
    \toprule
    \centering Datasets & \centering $\Delta R \%$(Ours) & \centering\arraybackslash $\Delta R \%$(R-D-$\lambda$ model\cite{RateControlLearned2022})\\
    \midrule
    \centering HEVC B & \centering \textbf{1.26} & \centering\arraybackslash 5.64\\
    \centering HEVC C & \centering \textbf{0.62} & \centering\arraybackslash 6.92\\
    \centering HEVC D & \centering \textbf{2.03} & \centering\arraybackslash 5.85\\
    \centering HEVC E & \centering \textbf{1.02} & \centering\arraybackslash 5.35\\
    \centering UVG & \centering \textbf{2.65} & \centering\arraybackslash 6.24\\
    \centering Average & \centering \textbf{1.66} & \centering\arraybackslash 6.05\\
  \bottomrule
  \end{tabular}
\end{table}

\noindent\textbf{Coding Complexity.}
In modern video transmission and processing, achieving real-time coding efficiency is crucial. Table \ref{tab:codingtime} displays our method's encoding and decoding times, showing an average encoding time of about 0.28 seconds per frame. For high-definition HEVC Class B (1080p) frames, this time is increased to less than 0.5 seconds due to higher computational needs. Notably, coding time is significantly reduced for low-bitrate routes (like Route 0) thanks to decreased computational complexity, showing the complexity optimization of our method. Table \ref{tab:complexity} compares coding complexities with DCVC-HEM, revealing our method's maximum MACs are 15\% lower than DCVC-HEM's. The RCA's computational load is marginal compared to the DRA, making our method's larger size a worthwhile trade-off for its superior rate control precision.

\begin{table}[t!]
  \caption{Encoding time and decoding time of proposed method on HEVC and UVG datasets.
  }
  \label{tab:codingtime}
  \centering
  \newcolumntype{C}{>{\centering\arraybackslash}X}
  \begin{tabularx}{\textwidth}{@{}*{1}{C}*{5}{C}@{}} 
    \toprule
    \multirow{2}{*}{Datasets} & \multicolumn{2}{c}{Route 0} & \multicolumn{2}{c}{Route 3}\\
    \cline{2-5}
    & \scriptsize Enc time (s/frame) & \scriptsize Dec time (s/frame) & \scriptsize Enc time (s/frame) & \scriptsize Dec time (s/frame)\\
    \midrule
    HEVC B & 0.312 & 0.276 & 0.435 & 0.29 \\
    HEVC C & 0.104 & 0.071 & 0.148 & 0.103 \\
    HEVC D & 0.050 & 0.039 & 0.063 & 0.051 \\
    HEVC E & 0.136 & 0.16 & 0.197 & 0.174 \\
    UVG & 0.318 & 0.282 & 0.376 & 0.305 \\
    Average & \textbf{0.216} & \textbf{0.192} & \textbf{0.280} & \textbf{0.212} \\
    \bottomrule
  \end{tabularx}
\end{table}

\begin{table}[t!]
  \caption{Complexity comparison between the proposed method and DCVC-HEM (1080p frame).}
  \label{tab:complexity}
  \centering
  \begin{tabular}{p{2.5cm}p{4cm}p{2.5cm}p{2.5cm}}
    \toprule
    \centering Methods & \centering MACs & \centering Coding Time (s) & \centering\arraybackslash Model Size (MB)\\
    \midrule
    \centering DCVC-HEM\cite{DCVC_HEM2022} & \centering 3.3T & \centering 0.781 & \centering\arraybackslash 67\\
    \centering Ours(Route 0) & \centering 2.1T(\textcolor{red}{DRA})+11.73G(\textcolor{blue}{RCA}) & \centering 0.569 & \multirow{4}{*}{\centering\makecell{61.02(\textcolor{red}{DRA})\\+25.84(\textcolor{blue}{RCA})}}\\
    \centering Ours(Route 1) & \centering 2.3T(\textcolor{red}{DRA})+11.73G(\textcolor{blue}{RCA}) & \centering 0.628 & \\
    \centering Ours(Route 2) & \centering 2.5T(\textcolor{red}{DRA})+11.73G(\textcolor{blue}{RCA}) & \centering 0.686 & \\
    \centering Ours(Route 3) & \centering 2.8T(\textcolor{red}{DRA})+11.73G(\textcolor{blue}{RCA}) & \centering 0.742 & \\
  \bottomrule
  \end{tabular}
\end{table}
 
\subsection{Ablation Study}
\label{subsec:ablation}
\begin{figure}[t!]
\centering
\includegraphics[width=120mm]{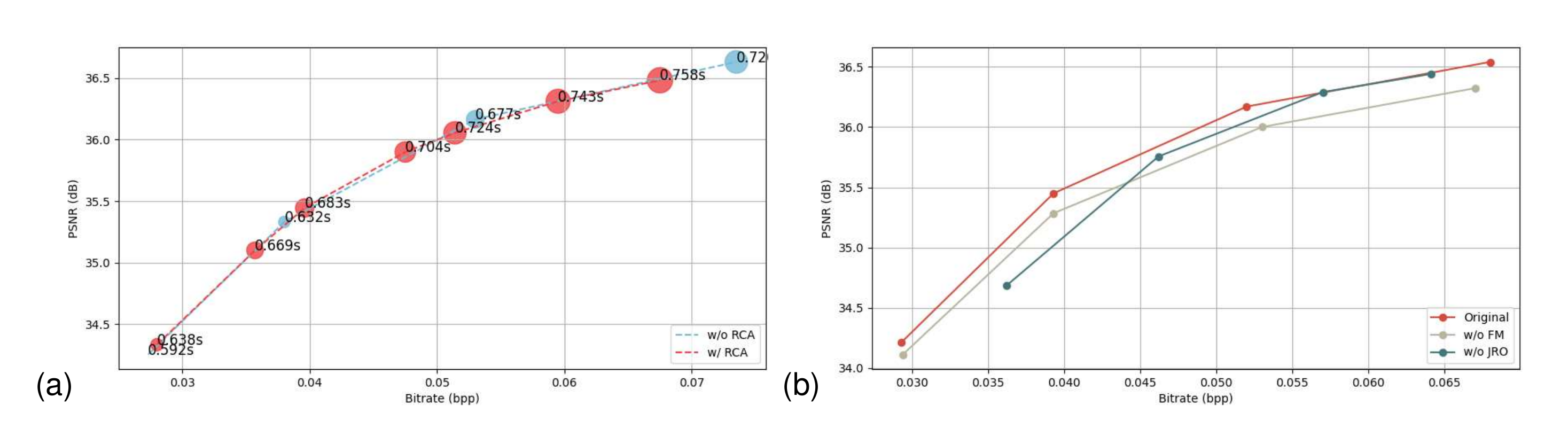}
\caption{Ablation study results on HEVC Class B. (a) Coding time comparison between the complete architecture and the architecture excluding RCA. (b) RD performance comparison: the original RD curve, performance without the Capacity Refinement Module, and without implementing the JRO strategy.}
\label{fig:ablation}
\end{figure}

\noindent\textbf{Impact of RCA on Coding Time}
Despite the lightweight RCA incurring minimal MACs compared to the DRA, this does not fully reflect on coding time due to parallel processing of GPUs. The evaluation on the HEVC Class B dataset reveals that incorporating the RCA into the DRA framework results in a marginal increase in coding time—merely around 0.05 seconds per frame, as shown in \cref{fig:ablation}. This insight underscores the proposed methods' capability to improve rate control accuracy without introducing significant time latency, making it a flexible solution for applications requiring both high coding efficiency and RD performance.

\noindent\textbf{Feature Modulation network and Joint-Routes Optimization}
The effects of the Feature Modulation network and the JRO strategy were explored by removing the former and substituting the latter with a basic training strategy (\cref{alg:naivetraining}). Results shown in \cref{fig:ablation} highlight two key insights: first, RD performance benefits from the FM network more significantly at higher bitrates due to the widening capacity gap between adjacent routes as the bitrate increases; second, excluding the JRO strategy leads to a noticeable decrease in RD performance at lower bitrates and narrows the spectrum of variable bitrates. These findings underscore the essential contributions of both the FM network and JRO strategy in enhancing RD efficiency across a spectrum of bitrates.

\section{Conclusion}
In this work, we propose a dynamic Neural Video Compression framework combining the DRA and the RCA for precise rate control and superior RD performance. This framework adaptively guides input frames through a series of coding routes, each defined by its complexity, to achieve a diverse range of bitrate. The lightweight RCA estimates bitrates based on frame content, facilitating adaptive rate control. Through extensive experiments, our framework exhibits remarkable RD performance and state-of-the-art precision in rate control, outperforming benchmark methods across various datasets. The proposed framework adaptively balances computational complexity with RD performance, offering a versatile solution for Rate-Distortion-Complexity Optimization in various bitrate and bitrate-constrained scenarios.



%
%
\bibliographystyle{splncs04}
\bibliography{main}
\end{document}